\documentclass[conference]{IEEEtran}
\IEEEoverridecommandlockouts
\usepackage{cite}
\usepackage{amsmath,amssymb,amsfonts}
\usepackage{algorithmic}
\usepackage{graphicx}
\usepackage{textcomp}
\usepackage{hyperref}
\usepackage{xcolor}

\usepackage{multirow}

\def\BibTeX{{\rm B\kern-.05em{\sc i\kern-.025em b}\kern-.08em
    T\kern-.1667em\lower.7ex\hbox{E}\kern-.125emX}}
\begin{document}

\title{PRSNet: A Masked Self-Supervised Learning Pedestrian Re-Identification Method}

\author{\IEEEauthorblockN{1\textsuperscript{st} Zhijie Xiao}
\IEEEauthorblockA{\textit{School of Information Science and} \\
\textit{Technology}\\
\textit{Tibet University}\\
Lhasa, China \\
JiayouMGei@163.com}
\and
\IEEEauthorblockN{2\textsuperscript{rd} Zhicheng Dong*}
\IEEEauthorblockA{\textit{School of Information Science and} \\
\textit{Technology}\\
\textit{Tibet University}\\
Lhasa, China \\
dongzc666@163.com}
\and
\IEEEauthorblockN{3\textsuperscript{nd} Hao Xiang}
\IEEEauthorblockA{\textit{Artificial Intelligence Academy } \\
\textit{Nanjing University of Aeronautics }\\
\textit{and Astronauticss} \\
Nanjing, China \\
xianghao@nuaa.edu.cn}
}

\maketitle

\begin{abstract}
In recent years, self-supervised learning has attracted widespread academic debate and addressed many of the key issues of computer vision. The present research focus is on how to construct a good agent task that allows for improved network learning of advanced semantic information on images so that model reasoning is accelerated during pre-training of the current task. In order to solve the problem that existing feature extraction networks are pre-trained on the ImageNet dataset and cannot extract the fine-grained information in pedestrian images well, and the existing pre-task of contrast self-supervised learning may destroy the original properties of pedestrian images, this paper designs a pre-task of mask reconstruction to obtain a pre-training model with strong robustness and uses it for the pedestrian re-identification task. The training optimization of the network is performed by improving the triplet loss based on the centroid, and the mask image is added as an additional sample to the loss calculation, so that the network can better cope with the pedestrian matching in practical applications after the training is completed. This method achieves about 5{\%} higher mAP on Marker1501 and CUHK03 data than existing self-supervised learning pedestrian re-identification methods, and about 1{\%} higher for Rank1, and ablation experiments are conducted to demonstrate the feasibility of this method. Our model code is located at \href{https://github.com/ZJieX/prsnet}{https://github.com/ZJieX/prsnet}.
\end{abstract}


\section{Introduction}
The term "pedestrian re-identification" was first introduced in 2005 by Wojciech Zajdel et al. from the University of Amsterdam \cite{zajdel2005keeping} in a study of multinodular visual tracking, and before the rise of convolutional feature extraction networks, researchers often used features such as pedestrian image color, texture, and shape \cite{matsukawa2016person} for their studies. With the continuous development of deep learning, more and more excellent pedestrian re-identification networks have appeared in public, from metric learning and representation learning in the early stages to unsupervised pedestrian re-identification algorithms nowadays.

Deep metric learning is the study of various loss functions to be used to enhance the robustness of the network. L. Zheng et al. \cite{zheng2017person} used pedestrian reclassification as an image multiclassification task based on ID loss, and improvements based on ID loss are often used on specific pedestrian reclassification tasks. Among them, many researchers have also proposed validation loss \cite{varior2016siamese}, \cite{deng2018image}, \cite{li2014deepreid}, \cite{zheng2017discriminatively} to optimize the relationship between 
\begin{figure}[!t]
\centerline{\includegraphics[width=1\linewidth]{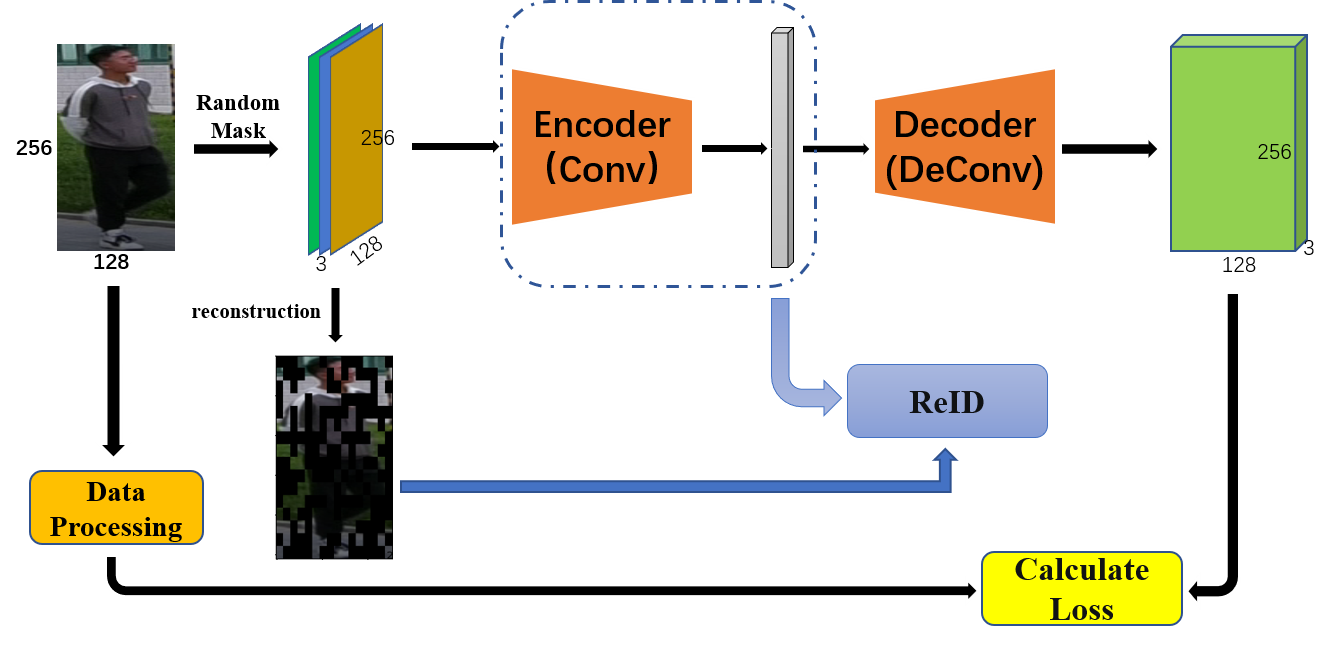}}
\caption{The structure of our masked self-supervised learning network framework, and we use a convolutional-deconvolutional network structure to form a autoencoder in self-supervised learning. The dashed box part of the figure is used to perform the pedestrian re-identification task after the self-supervised pre-training is completed.}
\label{fig1}
\end{figure}
positive and negative sample pairings, while validation loss is usually combined with ID loss \cite{chen2018group}, \cite{varior2016siamese} to improve the performance of the network; in 2015, Florian Schroff et al \cite{zheng2015scalable} proposed ternary group loss on a face recognition task, which can be trained to make the network since the original ternary loss is limited in discriminability. Hermans et al. \cite{hermans2017defense} proposed hard-sample ternary loss in 2017, i.e., in each training batch, the least similar positive samples and the most smaller negative samples are used for ternary loss training, which makes the network model very discriminative. Chen et al. \cite{chen2017beyond} proposed quadratic loss based on ternary loss, which makes the difference between positive and negative samples smaller and the difference between positive and negative samples larger, making the network more robust after training. These are the most popular deep metric learning approaches for pedestrian reidentification tasks. In representation learning, to capture fine-grained cues in global feature learning, a joint learning framework consisting of single image representation (SIR) and cross-image representation (CIR) \cite{wang2016joint} was studied early on, using specific sub-networks for triplet loss training. The widely used identity recognition embedding (IDE) model constructs the training process as a multiclassification problem by treating each identity as a distinct class. It is now widely used in pedestrian re-identification algorithms \cite{zheng2017unlabeled}, \cite{zhong2017re}, \cite{sun2018beyond}, \cite{sun2017svdnet}, \cite{ye2018robust}, and in 2016, Yutian Lin et al. \cite{lin2019improving} from the University of Technology Sydney proposed to use pedestrian attribute information to improve the network to recognize fine-grained feature information of pedestrians, which are classical global representation learning representation methods. For local representation learning, there are pedestrian re-identification methods that slice the image according to pedestrian body parts, such as PCB \cite{sun2018beyond}, AlignedReID \cite{zhang2017alignedreid}, and SCP \cite{fan2019scpnet}, and pedestrian re-identification methods based on human pose estimation, such as PIE \cite{zheng2019pose}, PDC \cite{su2017pose}, and GLAD \cite{wei2017glad}, as well as dividing the pedestrian image into many equidistant grids for local feature extraction, such as IDLA \cite{ahmed2015improved}, PersonNet \cite{li2018adversarial}, and classical networks such as DSR \cite{he2018deep}, where local features make it possible to resolve the spatial information that is not well taken into account in global features and which fuse all local features to obtain more detailed pedestrian features.

Although the above methods get good experimental results in the research, the results are not very good once they are applied, and the feature extraction network models of the above methods are pre-trained based on ImageNet datasets, which makes them have certain limitations in various practical applications, coupled with the single scene and low number of pedestrian re-identification datasets, so pedestrian re-identification in the last one or two years in the direction of unsupervised development. In 2020, Yixiao Ge et al. \cite{ge2020self} proposed SPCL as a cluster-based pseudo-label unsupervised learning algorithm, which provided a strong benchmark for unsupervised pedestrian reidentification. Based on this, Zuozhou Dai et al. \cite{dai2022cluster} proposed Cluster Contrast ReID, which improves the pseudo-label and training loss function based on this, which once surpassed the then supervised pedestrian reID algorithm by improving the pseudo-label and training loss function, and subsequently a large number of new pedestrian reID algorithms emerged according to the contract supervised learning in unsupervised learning, making the pedestrian reID method widely used in practical applications as well as achieving milestone results.

Based on the above-mentioned existing studies and problems, the main contribution points of the text are as follows:
\begin{itemize}
    \item[$\bullet$] Contrast learning applied to pedestrian recognition has a certain problem of confusing the discriminative attributes of person images. In the field of pedestrian re-recognition, whether it is a dataset or a pedestrian image keyed from other images or video images, there is a problem of small resolution, and the attribute information itself is fuzzy. Therefore, some prior tasks in contrast learning, such as image splitting and stitching, color transformation and image broadening recovery, may lead to the destruction of the target attributes in the pedestrian images, making the subsequent use of the pre-trained model for pedestrian re-identification tasks will produce certain interference, and not be able to perform well for pedestrian re-identification tasks. Therefore, this paper designs an image restoration-based generative learning to enable the network to dynamically extract features from important information areas of pedestrian images.
    \item[$\bullet$] Traditional feature extraction networks pre-trained by ImageNet are not able to fully exploit the fine-grained features of pedestrian images. According to self-supervised learning, we design a mask reconstruction proxy task instead of pre-training on ImageNet classification, which can make the network more suitable for the corresponding downstream tasks and also flexible to deal with the indirection problems caused by data and hardware.
    \item[$\bullet$] Also, in order to enable the network to focus precisely on the key regions in the pedestrian images, this paper puts the masked images in self-supervised learning to use, and we improve the centroid based triple loss to optimize the network training.
\end{itemize}

\section{Related Work}
In recent years, the self-supervised pedestrian re-identification algorithms have been subdivided into comparative learning and generative learning. Generative learning is represented by methods such as autoencoderss (e.g., GAN \cite{goodfellow2020generative}, VAE \cite{kingma2013auto}, etc.), which generate data from data so that it is similar to the training data in terms of overall or high-level semantics. In 2022, Kaiming He proposed MAE \cite{he2022masked}, which enables the network to easily cope with various computer vision downstream tasks by reconstructing important regions in images for pre-training. Zhongdao Wang et al. \cite{wang2020cycas} proposed that discriminative pedestrian re-identification features can be learned through the cyclic consistency of data association, which is an early attempt at self-supervised learning in the field of pedestrian re-identification. Hao Chen et al. \cite{chen2021joint} proposed to incorporate contrast learning and generative adversarial networks into a joint training framework to facilitate contrast learning between the original and generated pedestrian images. Zizheng Yao et al. \cite{yang2022unleashing} designed an unsupervised pre-training framework based on contrast learning to fully exploit the fine-grained local features in pedestrian images and enhance the global consistency between pedestrian images. Ke Han et al. \cite{han2022generalizable} investigated the generalization problem of pedestrian re-recognition and proposed the BNTA framework to address the fact that BN has a severe bias on the training domain, using a self-supervised learning strategy to adaptively update the BN parameters. Similarly, a good pedestrian re-identification algorithm is trained to require a reasonable loss function. Mikolaj Wieczorek et al. \cite{wieczorek2021unreasonable} propose to use the center of mass of classes in both the training and inference phases to alleviate the problems of computational speed and hard sample mining, where triple loss is optimized to represent each class by an embedding feature vector to accelerate the retrieval speed.

In this paper, we propose a self-supervised generative learning to complete the pre-training for the pedestrian re-identification task, and add the masked pedestrian image features to the ternary loss based on the center of mass so that the network in this paper can be more fully optimized for training.

\section{Proposed Method}
Generative learning can effectively reconstruct the features and information of the data itself, so it will not destroy the feature attributes in the pedestrian image, and there are certain problems of failure in discriminating the attributes of the task image, and the traditional feature extraction network pre-trained by the ImageNet dataset cannot fully explore the fine-grained problem of the pedestrian image. Based on the above, this paper proposes the self-supervised learning based on the generative mask and triple loss based on the mask center of mass to solve the above two problems, so that pedestrian re-recognition can be better applied in real life. The structure diagram of the proposed method is shown in Fig. 1. Only the encoder is kept for the pedestrian re-identification task, and the decoder will be discarded. The self-supervised pre-training method in this paper takes reference from MAE, but differs from it in that MAE uses two identical structures of ViT as encoder and decoder, while our method uses a symmetric convolution-deconvolution structure for the autoencoder. The method in this paper is described in detail below.

\subsection{Data Mask Preprocessing}
For this autoencoder, this section faces the pedestrian re-identification task, which requires an input image size of 128 × 256 × 3 (W × H × 3) for pre-training, and since pedestrian images are always irregular, the region blocks for masking are also irregular, and the pixel points of the region blocks after masking are replaced by 0 values. We define the width of each masked region divided as $p_w$ and the height as $p_h$, and $0<p_w<W$, $0<p_h<H$, and satisfy ${p_w}/{p_h}\ge2$. In this section, we choose to perform random 75{\%} region masking within the image region, see the reconstruction part in Figure 1, and the black part in the image is the masked part. Based on the above information, we perform a random mask on the pedestrian image, and then the masked part is recovered as much as possible after the autoencoder is trained.

\subsection{Autoencoder——Encoder}
When the data goes through the random mask, it first enters the encoder part, and the encoder structure is referred to as ConvNeXt \cite{liu2022convnet}, as shown in Fig. 2. 
\begin{figure}[ht]
\centerline{\includegraphics[width=1\linewidth]{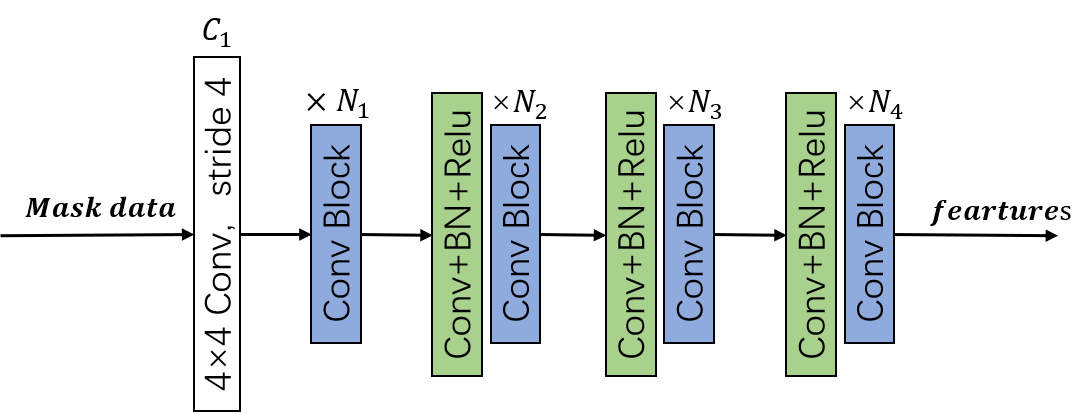}}
\caption{Encoder structure.}
\label{fig2}
\end{figure}
$N_1$, $N_2$, $N_3$ and $N_4$ represent the number of the network in different blocks, i.e., the network has a total of $N_1+N_2+N_3+N_4$ layers, and the larger these numbers are, the larger the number of parameters of the network is. In order to make the number of parameters of the network as small as possible, we use $N_1=3$, $N_2=3$, $N_3=9$ and $N_4=3$, so as to ensure the performance of the network while the number of parameter, the number of parameters is also not too large. We use the green structure in Fig. 2 to perform three downsampling operations, where the blue part only performs the convolution operation and does not change the size of the input feature layer each time. The size of the input mask data is required to be B × 3 × 256 × 128, and the size of the fearture obtained after the encoder is B × 2048, where the structure of each blue convolutional block is shown in (2) in Fig 3. Among them, in order to make the network learn the deep features of the image better, the multi-branched residual structure is carried out after each Conv Block, which is detailed in (1) in Fig. 3. 
\begin{figure}[ht]
\centerline{\includegraphics[width=1\linewidth]{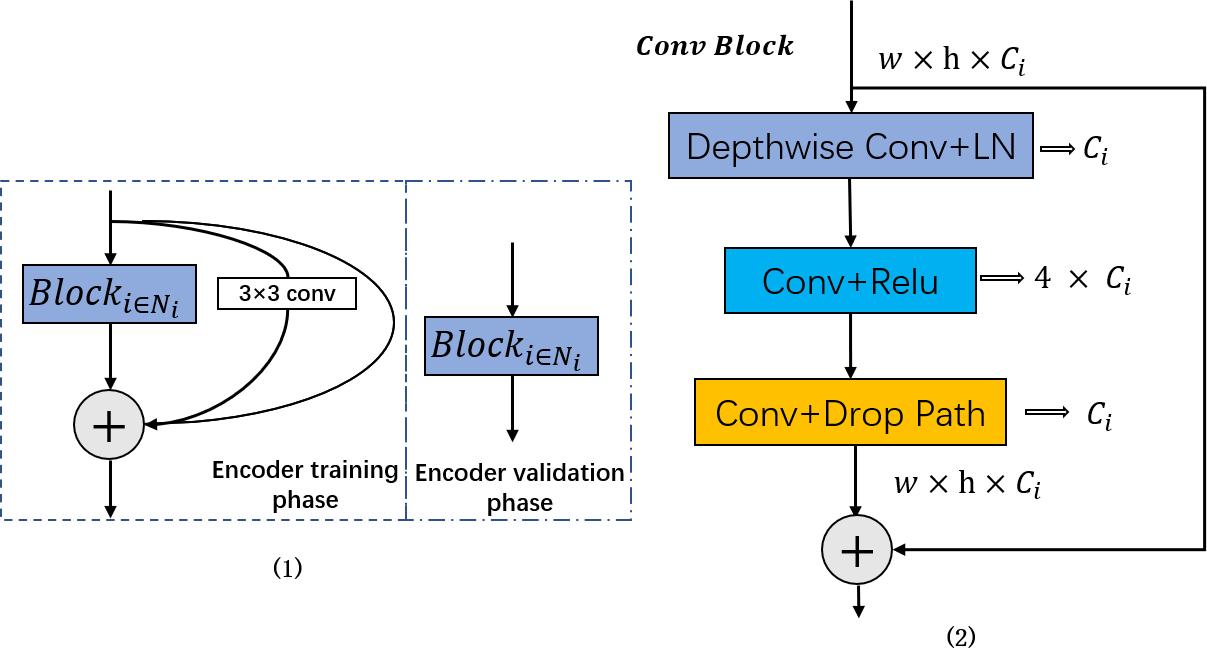}}
\caption{Training and verification strategy of Conv Block in network structure.}
\label{fig2}
\end{figure}
We refer to the structure of RepVGG \cite{ding2021repvgg} and only perform the residual operation during the training, and we discard the residual block in the test validation phase, because using the multi-branched residual block during the training can make the network learn the features well. In the validation phase, we discard the residual blocks because the use of multi-branched residual blocks during training enables the network to learn the image features well, but in fact, these parameters already exist in the network, and using the structural reparameterization, removing the residual structure in the validation phase will not degrade the network's performance, but will make the network occupy less video memory, and inference speed will be faster.

\subsection{Autoencoder——Decoder}
The decoder is a symmetric structure with the encoder, and its main purpose is to upsample the fearture to the same size as the original data. The structure of the decoder is shown in Fig 4, where $N_1$, $N_2$, $N_3$ and $N_4$ are the same as the values in the encoder, and the structure of the yellow part of the figure is also in shown Fig. 3 (2). The only difference between the decoder and the encoder is that we use deconvolution to upsample the features, and we do not use the RepVGG structure. When the feature is deconvolved with the first convolution kernel of size 8×4, a feature map of size B×C×8×4 will be obtained, and after that, it will be upsampled after each deconvolution, and the reconstructed data will be the same size as the original data.
\begin{figure}[ht]
\centerline{\includegraphics[width=1\linewidth]{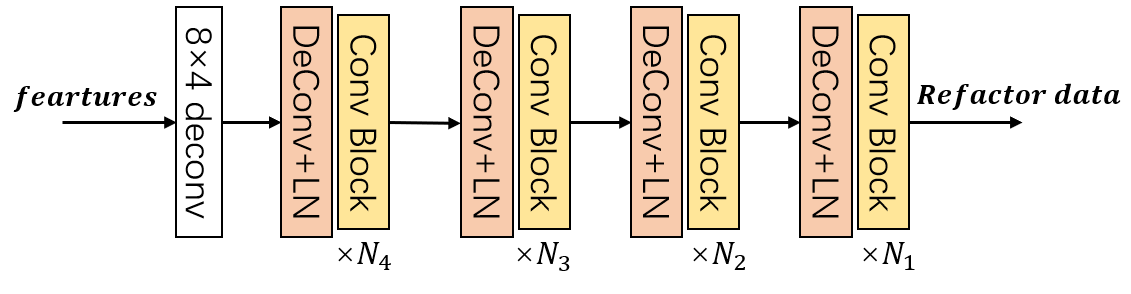}}
\caption{Decoder Structure.}
\label{fig2}
\end{figure}

\subsection{Loss Calculation}
The existing original data $X$, with $B$ original data in one training cycle, is obtained after a random mask operation as well as a mask sequence region m, with the value in m being 75{\%} of the image mask region location. When $X_mask$ is self-encoded to get the reconstructed data $X_rebuild$, the MSE loss function is used to optimize the training so that the difference between the reconstructed data and the original data in the random mask region is getting smaller and smaller, see (\ref{MSE LOSS}).
\begin{equation}\label{MSE LOSS}
    L = \frac{\sum_{i=0}^{len\left ( m \right)}{\left( X-X_{rebulid} \right) ^2\cdot m_i}}{B\cdot \sum_{i=0}^{len\left( m \right)}{m_i}}.
\end{equation}

\subsection{Triplet Loss Based on Masked Centroid}
After the above self-supervised pre-training, we remove the decoder and keep only the encoder structure for the pedestrian re-identification task, see Fig. 1. For the pedestrian re-identification task, we improve the triadic loss based on the center of mass, as detailed below.

The loss function proposed in this paper is an improvement on centroid-loss for the task of this paper. The triplet loss is given an anchor image $A$, and a positive sample image $P$ and a negative sample image $N$, and the objective is to minimize the distance between $A$ and $P$ while pushing the distance between $A$ and $N$. The detailed expressions are given in (\ref{tl}). Here $[x]_+=max(x,0)$ and $f$ represent the feature extraction network in the training phase, which is the encoder part of this paper.
\begin{equation}\label{tl}
    L_{tl}=\left[\left\| f\left(A \right) -f\left( P \right) \right| _{2}^{2}-\left\| f\left( N \right) \right\| _{2}^{2}+\alpha \right] _+.
\end{equation}

The triplet loss function based on the center of mass is to calculate the distance between the anchor image $A$ and its center of mass $c_P$ of all positive samples and the center of mass $c_N$ of all negative samples, where the center of mass is a simple average calculation of all the data, using sample center of mass aggregation This method produces a robust representation that is less susceptible to single-image mismatching and also reduces retrieval time because all positive and negative samples The same representation is already used for all positive and negative samples, so there is no need to compare each image. The expression of the centroid based triplet loss function (\ref{ctl}).
\begin{equation}\label{ctl}
    L_{ctl}=\left[ \left\| f\left( A \right) -c_P \right\| _{2}^{2}-\left\| f \left( A \right) -c_N\right\| _{2}^{2}+\alpha _c \right] _+.
\end{equation}

In this paper, the pedestrian image data will be randomly masked before entering the autoencoder, so we reconstruct the masked pedestrian image and input it into the ternary loss based on the center of mass as part of the data; see Fig. 1. Based on this, we propose a ternary loss function based on the masked center of mass, as shown in eq.(\ref{mctl}), where the masked image and the original image labels are all the same, the difference being that the original image is The difference is that the original image is intact, while the image after random masking is missing many regions, which is neither a complete image, where $m_P$ and $m_N$ are the centers of mass of all positive and negative samples of the masked image, and $\lambda _1+ \lambda _2=1$, which is based on the ratio of the number of positive and negative samples in the masked image.
\begin{equation}\label{mctl}
\begin{aligned}
    L_{mctl}=\left[ \left\| f\left( A \right) -\lambda _1m_P-c_P \right\| _{2}^{2} \right. -\\ \left.
    \left\| f\left( A \right) -\lambda _2m_N-c_N \right\| _{2}^{2}+\alpha \right] _+.
\end{aligned}
\end{equation}
With this loss function, the network can actively focus on the important regions of the pedestrian images and learn the fine-grained information of the images while having the advantages of the centroid based triple loss function.

\section{Experiment}
In this paper, we first perform generative self-supervised learning for pre-training, i.e., the original image is passed through a random mask, and then the masked part is reconstructed after the autoencoder training, and then the decoder is removed from the autoencoder, in which the pre-trained model weights are used to load into the encoder, which performs the pedestrian re-recognition task.

The data set used for pre-training is the COCO2017 data set and the Pascal VOC data set with the category of human images, and the human is cropped-out using its annotation information, and the cropped out pedestrians are keypoint detected using the human pose estimation network HRNet \cite{sun2019deep}, and the pedestrian images with incomplete keypoints are removed. At this time, the data set does not have any labels, and the training set contains a total of 80689 pedestrian images, and these image data are mainly used for self-supervised learning. For the pedestrian re-identification task, we use four classical datasets: Market151 \cite{zheng2015person}, CUHK03 \cite{li2014deepreid}, MSMT17 \cite{wei2018person}, and Dukemtmc-Reid \cite{zheng2017unlabeled} for training and evaluation respectively.

\begin{table}[htbp]
\renewcommand{\arraystretch}{1.5}
\caption{Experimental environment}
\begin{center}
\begin{tabular}{ccc}
\hline 
1 & cpu & Intel(R) Xeon(R) Platinum 8160 \\
\hline
2 & memory & 15G \\
\hline
3 & operating system & CentOS7 \\
\hline
4 & video card & NVIDIA TESLA T4 × 8 \\
\hline
\end{tabular}
\label{tab1}
\end{center}
\end{table}

The methods in this paper are used to compare with the SOAT network on the pedestrian re-identification task by self-supervised learning in recent years. Also, since the methods in this section refer to MAE, a comparison test is done between the methods in this section and MAE using the same training method. We also compare the ternary group loss of mask prime proposed in this chapter with centroid-loss on four pedestrian reidentification datasets for experimental comparison. The experimental environment of this paper is shown in Table \uppercase\expandafter{\romannumeral1}, and the detailed experimental comparison is as follows.

\subsection{Pre-training Visualization Results}
We use the model with the lowest loss value in pre-training for the visualization of the pre-training part, and we will use this model to initialize the weight parameters of the encoder for the pedestrian re-identification task. The test dataset we use is the pedestrian images taken by ourselves without any correlation with the training data, and the detailed results are shown in Fig. 5. 
\begin{figure}[ht]
\centerline{\includegraphics[width=1\linewidth]{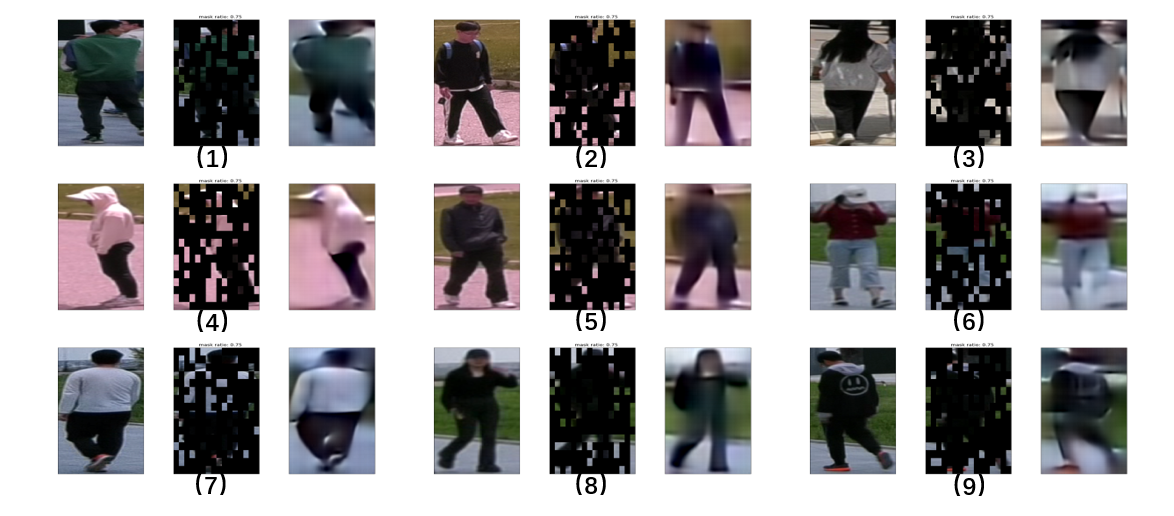}}
\caption{Masked self-supervised learning for visualization of pedestrian image reconstruction results.}
\label{fig3}
\end{figure}
Nine randomly selected images are shown among all the test results, where the leftmost image on the way shows the original image, the middle image shows the image obtained after the 75{\%} random mask operation, and the right image is the image reconstructed by the autoencoder. The figure shows that some details in the pedestrian image are still not well restored, probably because the convolutional structure in the decoder cannot learn the original content of the image well due to the low resolution of the pedestrian image, and we will introduce the content of GAN to improve this part of the pedestrian image reconstruction to make the reconstruction more detailed in the future.

At the same time, in order to visualize the image feature extraction by the network more intuitively, we also performed a heat map operation on the masked image dropped into the network, as shown in Fig. 6. 
\begin{figure}[ht]
\centerline{\includegraphics[width=1\linewidth]{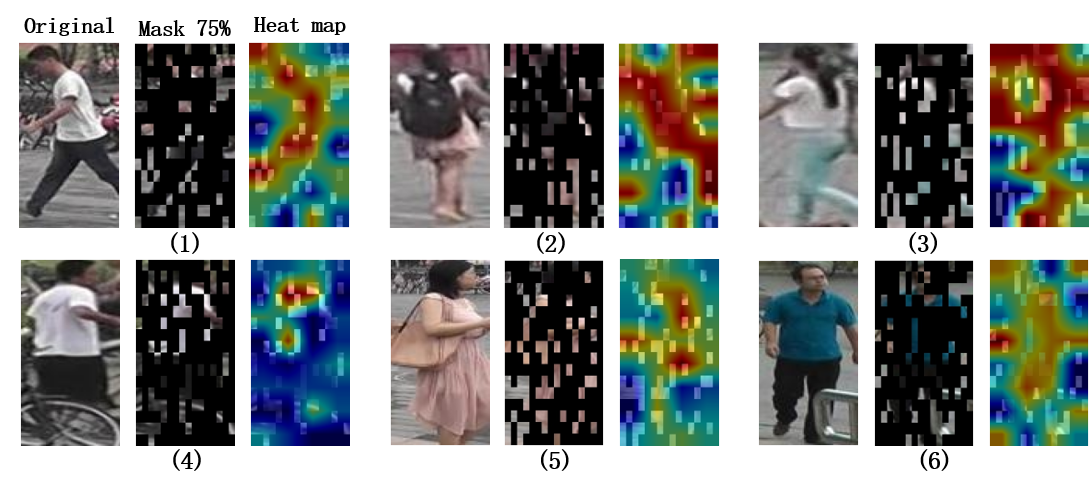}}
\caption{Results of network visualization of heat map of           masked images.}
\label{fig4}
\end{figure}
It can be seen that the network can still focus well on the general area of the pedestrian in the image despite the image being masked, which also proves that the proposed method in this paper can better focus on the fine-grained information of the pedestrian image after the masked self-supervised learning training.

\subsection{Experimental Results of Pedestrian Re-identification}
As shown in Fig. 1, after the masked self-supervised learning training, we just used the encoder for the pedestrian re-identification task, and performed the training as well as the evaluation on top of four classical datasets. We then compared the most advanced self-supervised learning methods in recent years on the Market1501 and CUHK03 datasets, and the detailed evaluation results are shown in Table \uppercase\expandafter{\romannumeral2}. mAP obtained in this section is more than 5 points higher than the existing SOAT self-supervised learning methods on these two datasets. The mAP obtained in this section is more than 5 points higher than the existing SOAT self-supervised learning methods on these two datasets, but the only shortcoming is that the Rank1 index does not get much improvement and is basically the same as it is. In pedestrian re-recognition, mAP is a response to the degree that all correct images in the retrieved human hits are ranked at the top of the sorted list, which is used to evaluate the overall effectiveness of the pedestrian re-recognition method and can measure the performance of a pedestrian re-recognition method more comprehensively.
\begin{table}[htbp]
\renewcommand{\arraystretch}{1.5}
\caption{The results of comparing the method in this          paper with the advanced self-supervised learning          methods in recent years}
\begin{center}
\begin{tabular}{c|cc|cc}
\hline
 \multirow{2}{*}{\textbf{Methods}} & \multicolumn{2}{c}{\textbf{Market1501}} & \multicolumn{2}{|c}{\textbf{CUHK03}} \\
\cline{2-5}
 & mAP(\%) & Rank1(\%) & mAP(\%) & Rank1(\%) \\
\hline 
PCB & 81.6 & 93.8 & 57.5 & 63.7  \\
OSNet & 84.9 & 94.8 & 67.8 & 72.3 \\
P2Net & 85.6 & 95.2 & 73.6 & 78.3 \\ 
DSA & 87.6 & 95.7 & 75.2 & 78.9 \\
GCP & 88.9 & 95.2 & 75.6 & 77.9 \\ 
SAN & 88.0 & 96.1 & 76.4 & 80.1 \\
ISP & 88.6 & 95.3 & 74.1 & 76.5 \\
GASM & 84.7 & 95.3 & - & - \\
RGA-SC & 88.4 & 96.1 & 77.4 & 81.1 \\
HOReID & 84.9 & 94.2 & - & -  \\
AMD & 87.1 & 94.8 & - & - \\
TransReID & 89.5 & 95.2 & - & - \\
PAT & 88.0 & 95.4 & - & - \\
MGN+UP-ReID & 91.1 & \textbf{97.1} & 85.3 & 87.6 \\
\hline
ours & \textbf{97.6} & \textbf{97.1} & \textbf{90.8} & \textbf{88.1} \\
\hline
\end{tabular}
\label{tab1}
\end{center}
\end{table}

Meanwhile, to better prove the reliability of the method proposed in this chapter, we conducted peer-to-peer experiments using MAE, i.e., we first performed masked self-supervised learning pre-training, and then selected the one with the lowest training loss model for the pedestrian re-identification task. The resulting data are shown in Table \uppercase\expandafter{\romannumeral3}, from which it can be seen that in each of the four datasets, our method is higher than MAE by more than 3 points in each metric. This also proves that the method in this section is indeed effective.
\begin{table}[htbp]
\renewcommand{\arraystretch}{1.5}
\caption{Comparison of the evaluation of the method in this paper with MAE after equivalent training conditions}
\begin{center}
\begin{tabular}{c|cc|cc}
\hline
 \multirow{2}{*}{\textbf{Methods}} & \multicolumn{2}{c}{\textbf{Market1501}} & \multicolumn{2}{|c}{\textbf{CUHK03}} \\
\cline{2-5}
 & mAP(\%) & Rank1(\%) & mAP(\%) & Rank1(\%) \\
\hline 
MAE & 92.4 & 90.8 & 87.0 & 82.6 \\

ours & \textbf{97.6} & \textbf{97.1} & \textbf{90.8} & \textbf{88.1} \\
\hline
\multirow{4}{*}{} & \multicolumn{2}{c}{\textbf{DukeMTMC}} & \multicolumn{2}{|c}{\textbf{MSMT17}} \\
\cline{2-5}
& mAP(\%) & Rank1(\%) & mAP(\%) & Rank1(\%) \\
\cline{2-5}
& 89.5 & 90.2 & 71.9 & 65.9 \\ 

& \textbf{95.5} & \textbf{94.7} & \textbf{86.4} & \textbf{83.4} \\
\cline{2-5}
\end{tabular}
\label{tab1}
\end{center}
\end{table}

We also conducted an experimental comparison using centroid-loss (ctl) and the mask-centroid-loss (mctl) proposed in this chapter on four datasets; see Table \uppercase\expandafter{\romannumeral4}, which shows that, according to the data, the metrics obtained after training using the loss function proposed in this chapter are both higher than those obtained by centroid-loss by at least 0.2 points. This also reflects that our proposed mctl enables the network to learn more potential features in pedestrian images.
\begin{table}[htbp]
\renewcommand{\arraystretch}{1.5}
\caption{Comparison of post-training evaluation using ctl and mctl, respectively}
\begin{center}
\begin{tabular}{c|cc|cc}
\hline
 \multirow{2}{*}{\textbf{Methods}} & \multicolumn{2}{c}{\textbf{Market1501}} & \multicolumn{2}{|c}{\textbf{CUHK03}} \\
\cline{2-5}
 & mAP(\%) & Rank1(\%) & mAP(\%) & Rank1(\%) \\
\hline 
ours(ctl) & 97.4 & 97.0 & 90.3 & 87.2  \\

ours(mctl) & \textbf{97.6} & \textbf{97.1} & \textbf{90.8} & \textbf{88.1} \\
\hline
\multirow{4}{*}{} & \multicolumn{2}{c}{\textbf{DukeMTMC}} & \multicolumn{2}{|c}{\textbf{MSMT17}} \\
\cline{2-5}
& mAP(\%) & Rank1(\%) & mAP(\%) & Rank1(\%) \\
\cline{2-5}
& 95.3 & 94.3 & 86.2 & 83.1 \\ 

& \textbf{95.5} & \textbf{94.7} & \textbf{86.4} & \textbf{83.4} \\
\cline{2-5}
\end{tabular}
\label{tab1}
\end{center}
\end{table}

Finally, we partially visualize the evaluation results on the Market1501 dataset; see Fig. 7, where three pedestrians with different ids are selected as visualization results, where column 1 is the query image of different perspectives of the same pedestrian, and the next 10 columns are matched as the same target, sorted by similarity from largest to smallest, and the network is basically the same target for all matching
results within Rank10.

\section{Conclusions}
In this paper, we propose a generative self-supervised learning method to address the problem of confusing discriminative attributes of human images and the problem that traditional feature extraction networks pre-trained by ImageNet cannot fully exploit the fine-grained nature of human images when applied to the pedestrian re-recognition task, and propose a masked After training and evaluation on four classical pedestrian re-recognition datasets, the proposed method achieves good results in comparison with existing SOAT models and loss function experiments, which also proves the robustness of the method in this chapter.

\end{document}